\documentclass[conference]{IEEEtran}

\ifCLASSINFOpdf
\else
\fi

\usepackage{graphicx}
\graphicspath{ {./pics/} }
\usepackage{tikz}
\usetikzlibrary{shapes,arrows}
\usepackage{caption}

\usepackage[utf8]{inputenc}
\usepackage{tabularx}
\newcolumntype{L}{>{\raggedright\arraybackslash}X}
\usepackage{makecell}
\usepackage{booktabs}
\usepackage[style=numeric,citestyle=ieee, sorting=none]{biblatex}
\usepackage{adjustbox}
\usepackage{lineno,hyperref}
\usepackage[ruled,vlined]{algorithm2e}
\usepackage{amsmath}
\usepackage{adjustbox}
\usepackage{lscape}
\usepackage{multirow}
\SetKwInOut{Parameter}{Parameters}

\usepackage{subcaption}

\usepackage{amssymb}

\begin{document}

\title{On Principal Curve-Based Classifiers and Similarity-Based Selective Sampling in Time-Series}

\author{\IEEEauthorblockN{Aref~Hakimzadeh}
\IEEEauthorblockA{School of Electrical and \\Computer Engineering\\
Shiraz University\\
Shiraz, Fars\\
Email: sa.hakimzadeh@shirazu.ac.ir}
\and
\IEEEauthorblockN{Koorush~Ziarati}
\IEEEauthorblockA{School of Electrical and \\Computer Engineering\\
Shiraz University\\
Shiraz, Fars\\
Email: ziarati@shirazu.ac.ir}
\and
\IEEEauthorblockN{Mohammad~Taheri}
\IEEEauthorblockA{School of Electrical and \\Computer Engineering\\
Shiraz University\\
Shiraz, Fars\\
Email: mtaheri@cse.shirazu.ac.ir}
}

\maketitle

\begin{abstract}
Considering the concept of time-dilation, there exist some major issues with recurrent neural Architectures. Any variation in time spans between input data points causes performance attenuation in recurrent neural network architectures. Principal curve-based classifiers have the ability of handling any kind of variation in time spans. In other words, principal curve-based classifiers preserve the relativity of time while neural network architecture violates this property of time. On the other hand, considering the labeling costs and problems in online monitoring devices, there should be an algorithm that finds the data points which knowing their labels will cause in better performance of the classifier. Current selective sampling algorithms have lack of reliability due to the randomness of the proposed algorithms.
This paper proposes a classifier and also a deterministic selective sampling algorithm with the same computational steps, both by use of principal curve as their building block in model definition.

\end{abstract}

\section{Introduction}
\label{intro}
Label prediction in time-series is equivalent to event recognition. A time-series classifier should be able to recognize events instead of trend. A particular event can happen in different time spans; which means two events with the same label may occur in different time duration. Classifiers that are based on neural networks, e.g., recurrent neural networks (RNN), lack the ability of event recognition in a manipulated time stamp. In other words, RNNs are particular solutions to the event recognition problem with a specified time-span.
\par
\textbf{The ability to recognize an event is the ability to detect
that event in several time steps.}
Several sources endorse this statement, as Jeffrey M.Zacks says, \emph{"From a range of sources comes evidence that people perceive and conceive of events in terms of representations that span time-scales", "People spontaneously segment activity at different
timescales in correspondence with a partonomic hierarchy"(Jeffrey M.Zacks, 2001, Psychological bulletin)}. He later mentioned that, \emph{"The theory asserts that people do not perceive event boundaries on only one timescale. Rather, they perceive event boundaries on multiple
timescales simultaneously but selectively attend to one timescale in response to instructions or other experimental manipulations" (Jeffrey M.Zacks, 2007, Psychological bulletin)} 
\par
For instance, if a child is able to count integer numbers, he would also be able to count with time step 2 (counting just even or odd numbers). Also, if a child recognizes a specific word in English, he can also recognize it with a different accent. This problem includes any variation in the time parameter and not just a regular change in the sampling rate. An example in the AI world can be a device which is not real time and sends sampled data in irregular intervals. How  this problem affects prediction systems and how we can handle it, are the subjects of current paper.
\par
In many real-world problems, human should do the labeling as a costly task. In this case, an algorithm to decide which data points are more worthy to for labeling can be beneficial. In other words, considering a limited budget for querying label of a data point, it is preferred to query the labels of the points that will cause better performance of the classifier. This problem is known as selective sampling and the budget for querying the label of a data point is known as the query budget. Selective sampling has similarities with anomaly detection problem.
\par
In this paper, recurrent and principal curve-based classifier have been compared. Also, a deterministic selective sampling method is proposed with an ignorable time consumption in comparison with the task of label prediction.
\section{Literature Review}
In this paper, the literature review is partitioned into methods based on principal curves, recurrent structures in neural networks and selective sampling. In each part several prominent algorithms in that field are reviewed.
\subsection{Principal Curves}
Principal curve was defined to be a non-parametric curve that passes through the cloud of data \cite{hastie1989principal}. More specifically, each point in a principal curve is the expected value of the data points that have been projected on that specific point. The original objective function is hard to reach and applicable implementations of principal curves try to find a sub-optimum solution. One of the promising modifications of the original objective function breaks the curve to K-1 linear forms \cite{kegl1999polygonal} separated by K points, which represent the model of data cloud. Each of the K-1 lines will be optimized considering all data points that have been projected on that specific line-segment. However, the K principal curve accomplished in fitting the curve to different data distributions, is much time consuming because the algorithm surfs all data points in each step of optimization. To address time consumption problem in principal curves, Constraint Local Principal Curve (CLPC) method proposed an algorithm which excludes a batch of data points from optimization process in each step \cite{chen2016constraint}. CLPC method determines the first and last data points as the first and last principal curve points. Then, in each step, it appends a curve point in between until the total error does not exceed a predefined threshold. Assuming that $P_{i}$ is the $i_{th}$ point of the extracted principal curve, the first and second data points are considered as $P_{1}$ and $P_{K}$ respectively. For appending each new point in the curve, a circle with $P_{n-1}$ as the center and radius equal to $\|P_{n}-P_{n-1}\|$ is drawn. The expected value of the data points in a narrow bar around drawn circle’s perimeter will be enacted as the new curve point if the error is less than the predefined threshold. After determining each curve point, all the data points that have been projected on the former parts of the curve will not participate in future calculations.
As far as we know, there are few papers on classifiers using principal curves. A classifier was proposed in \cite{qi2010microarray} that trains one principal curve for each class of data. Then in the test phase, each sample is assigned to its nearest principal curve and corresponding class. That method is not able to handle sequential data.
\subsection{Recurrent Structures}
The pioneer of prediction in time-series includes Recurrent Neural Networks(RNN). These structures are meant to preserve the effect of each data point in later time steps by passing a history unit through the network flow. Recurrent structures have shown promising results in many real-world problems like natural language processing, genomic analysis, image captioning and etc. \cite{auli2013joint, lipton2015learning, vinyals2015show}. Early efforts included supervised learning on sequential data with simple recurrent structures \cite{elman1990finding}. However, RNN structures were unable to preserve long term effects and lacked the long-term memory characteristics. To address this problem, Long-Short-Term-Memory(LSTM) cells were proposed to handle both long and short-term effects \cite{hochreiter1997long}. Adding forget gates to LSTM cells made them more powerful memory units for predicting sequential data\cite{gers1999learning}. LSTM cells have excessive parameters which demands numerous data points for training the network. Another proposal for preserving long and short-term effects, and handling vanishing gradients, is Gated Recurrent Units(GRU) by adding reset and forget gates to RNN structures\cite{cho2014learning}. Compared to LSTM cells, GRU was simpler and has fewer parameters. Another approach to improve the generalization capacity of recurrent structures is dropout \cite{pham2014dropout}. Some other approaches tried to implement a convolutional layer on top of LSTM cells \cite{karim2017lstm} which paved the way to implement attention mechanisms based on recurrent structures\cite{karim2019insights}. Affected by deep learning, deep RNN was introduced to be able to learn more complex dependencies between data points by adding recurrent layers on top of each other\cite{pascanu2013construct}. 

Non-recurrent models like Gaussian models with latent variables are some other solutions for sequential data analysis\cite{ghassemi2015multivariate, stanculescu2014hierarchical}. However, recurrent structures have better performance in preserving long-term effects. Temporal Convolutional Network (TCN) is a neural network architecture that is able to preserve long-term dependencies in sequential data \cite{oord2016wavenet}. Unlike the former methods, TCN is not a recurrent structure but has achieved promising performance in voice recognition. The main drawback of RNN models in event prediction is their assumption on having data points with equal time steps.

\subsection{Selective Sampling}
Selective sampling is about finding data points that knowing their label will cause better training for a specific classifier. 
A usual approach in selective sampling is finding distribution or decision boundaries which determine the corresponding regions of the classes. Each input data point belongs to each class based on the defined distribution or the distance to the boundaries \cite{cesa2003learning}. Most of the selective sampling algorithms, sample from data points by a Bernoulli distribution extracted from the class probabilities. For example, in perceptron selective sampling, the class probabilities are determined by one or more perceptron(s) to determine the selection threshold for Bernoulli sampling\cite{cesa2006worst}. Winnow and second-order-perceptron selective sampling algorithms use the same query strategy with a different method of probability estimation\cite{cesa2006worst}. Some of the sampling algorithms are known as active learning, where a Gaussian distribution is used to model the class probabilities. The generated output of the Gaussian distribution is used to compute the query probability of the corresponding data point and sampling from Bernoulli distribution with inferred threshold makes the query decision\cite{sheng2008get}. Considering imbalanced datasets, Cost-Sensitive Online Active Learning(CSOAL) was proposed to assume different probability distributions for different classes of data\cite{zhao2013cost}. Defining an asymmetric update rule for each class of data led to Online Asymmetric Active Learning(OAAL) algorithm\cite{zhang2016online}. Combining the asymmetric update rule and asymmetric query probability was proposed in \cite{zhang2018online}. Online Adaptive Asymmetric Active Learning(OA3L) was designed for a binary-class imbalanced data where two distinct Bernoulli distributions with different hyper-parameters make the query decision considering the computed probability and asymmetric update rules updates the boundaries of each class of data differently.
\par
Selective sampling is, somehow, finding the most informative data points in the training phase. Some proposed algorithms try to rank the data points considering an uncertainty evaluation. For example, in \cite{smailagic2020medal} data points with highest entropy are found and the most distant samples among them are recommended as queries. Information theoretical approaches form another set of selective sampling methods; but most of them are not designed to sample in online manner. Furthermore, these methods do not preserve the sequential characteristics of the data.
\par
As far as we know, all selective sampling algorithms have a randomness inside the algorithm which causes unreliability of the sampling algorithm in applications with high sensitivity. Also, randomness leads to more misclassification for imbalanced datasets.
\section{Proposed Algorithm}
The following paragraph summarizes the notation used in this paper. In the following sections, it is considered that there are $n$ training data points $\{ X_{1},X_{2},…,X_{n}\}$. The $i_{th}$ training data point is denoted by an ordered pair $X_{i}=(t_{i},x_{i})$ where $t_{i}$ is the time of observing the data point $x_{i} \in \mathbb{R}^{d}$ with $d$ features. Also,  $P_{k}$ is the $k^{th}$ point of the principal curves assuming that the principal curves of all classes form a unit principal curve(by concatenating extracted principal curves). The stated united principal curve is called \emph{Governing pattern} in this paper. It is notable that the $P_{k,0}$ represents the time variable of the $k_{th}$ principal curve point($P_{k}$). Parameter $H$ is the history length of the classifier which determines the length of the input stream stored for predicting the next input. In other words, in each step, classifier uses last $H$ input data points to predict the label of the current input point. Parameter $C$ is the number of classes in the input data. In addition, $t_{o}$ is the time shift(offset) of the input data stream in the matching step. $X_{i,t_{o}}$ is used to show that $X_{i}$ is shifted in time with value of $t_{o}$. In case of projection on the principal curves, $f_{\tau}(.)$ denotes the projection on the $\tau^{th}$ principal curve(corresponding to the $\tau^{th}$ class). Also, $F(.)$ is the projection on the nearest principal curve, Eq. \eqref{eqn:projrel} shows the stated relation.
\begin{equation}\label{eqn:projrel}
	\begin{aligned}
	& ~~~~~~~~~~~~~~~ F(X_{n,t_{o}}) = f_{\beta}(X_{n,t_{o}}) \\
	& where ~~ \beta = \underset{\tau}{argmin} ~ \| f_{\tau}(X_{n,t_{o}})- X_{n,t_{o}} \|
	\end{aligned}
\end{equation}
\par
This section consists of two parts. In the first part, the modeling and optimization steps of the proposed classifier is explained and in the second part, the extra step for selective sampling is explained. Both the proposed classifier and sampling algorithms have common steps for initialization, matching process and update rules. Fig. \ref{fig:Algorithm} demonstrates the flowchart of the proposed algorithm.

\tikzstyle{decision} = [diamond, draw, fill=blue!20, 
    text width=4.5em, text badly centered, node distance=3cm, inner sep=0pt]
\tikzstyle{block} = [rectangle, draw, fill=blue!2, 
    text width=5em, text centered, rounded corners, minimum height=4em]
\tikzstyle{pic_block} = [rectangle, draw, rounded corners,
    minimum height= 4cm,minimum width= 4cm]
\tikzstyle{line} = [draw, -latex',line width=1mm]
\tikzstyle{dline} = [draw, -latex', fill=gray!10,line width=1mm]
\tikzstyle{cloud} = [draw, ellipse,fill=gray!20, node distance=3cm,text width=5em,minimum height=2em]
\tikzstyle{dec} = [draw, diamond,fill=gray!20,minimum height=1cm, minimum width=2cm]
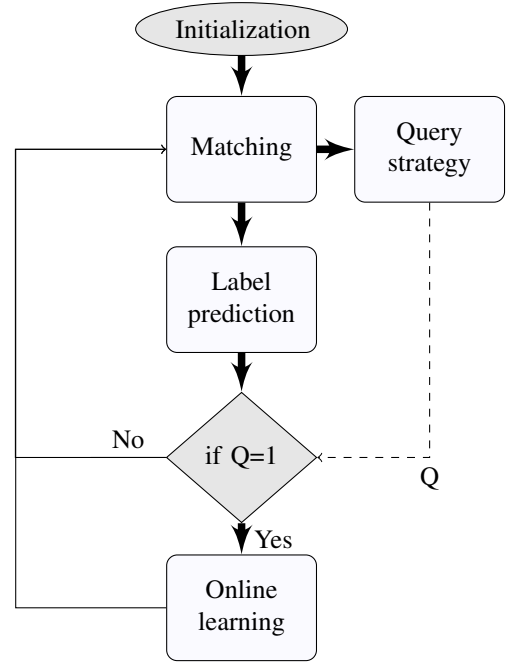
\begin{figure}
\caption{\label{fig:Algorithm}Flowchart of the proposed classifier and the proposed selective sampling algorithm. }
\begin{center} 
\begin{tikzpicture}[auto]
    \node [cloud, node distance=1cm] (init) {Initialization};

    \node [block, below of=init, node distance=1.6cm] (matching) {Matching};
    \node [block, right of=matching, node distance=2.5cm] (query) {Query strategy};
    \node [block, below of=matching, node distance=2cm] (predict){Label prediction};
    \node [dec, below of=predict, node distance=2.1cm] (iff) {if Q=1};    
    
    \node [block, below of=iff, node distance=2cm] (learning) {Online learning};
    \path [dline] (init) -- (matching);
    \path [line] (matching) -- (predict);
    \path [line] (matching) -- (query);
    \path [line] (predict) --  (iff);
    \path [line] (iff) -- node {Yes} (learning);
    \draw[->,dashed] (query.south) |- node {Q} (iff.east);
    \draw[->] (iff.west) -- ++(-2,0) |-  (matching.west);
   \draw (iff.west) -- ++(-1,0) node[midway,above]{No};
    \draw[->] (learning.west) -- ++(-2,0) |- (matching.west);

\end{tikzpicture}
\end{center}
\end{figure}

As shown in Fig. \ref{fig:Algorithm}, after modeling the input data in \emph{initialization} step, \emph{matching} step finds the most similar state between the input data stream and the \emph{governing pattern}. \emph{Label prediction} step finds the nearest curve to the matched data and \emph{query strategy} decides the query considering the similarity history. If query strategy decides to ask the true label, \emph{online learning} will enhance \emph{governing pattern} with queried label.
\subsection{Classifying}
In this section, the proposed classifier is explained in details. At the initialization step, the input data points are modeled with principal curves and then the \emph{governing pattern} is extracted. This type of modeling considers the time as an informative feature and also preserves relativity characteristics of time in all subsequent steps. Matching process is the next step to search for a time shift with the minimum projection error. It is also able to handle the variant time spans between the data points. In the next step, the classifier predicts the label of the input data as the corresponding label of the matched input stream. In the last step, the classifier enhances the extracted model considering the true label of the data and the predicted label.
\subsubsection{Initialization}
In initialization step, algorithm is meant to extract a model from an initial labeled dataset. In the proposed method, it is desired to fit several K-principal curves to data, one curve per class. Obviously, each point is presented with the same number of features as the input data. CLPC method is used, in this paper, for extracting the principal curves. In addition, after finding all curve points with CLPC method, the curve points with angle (between two connected line-segments) less than a predefined threshold are removed. This reform in curve points causes removing less informative points and reducing computational cost in \emph{matching} step.
\begin{figure}[h]
	\caption{\label{fig:init} Initialization step. Each color defines a class of data and pale points are the original data points.}
    \center{\includegraphics[width=\linewidth]{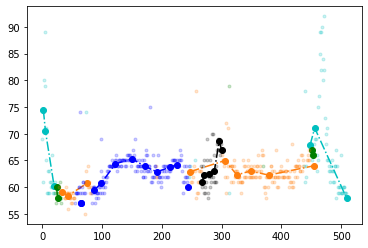}}    
\end{figure}
\newline
Fig. \ref{fig:init} shows this step applied on one feature of "Sleep" dataset. Pale points are the input points and each class is represented by a different color. Therefore, the proposed model has $K \times (d+1)$ parameters where $K$ is the number of points in the curve and $d$ is the feature dimension of data. In this paper, the extracted principal curve is called as \emph{governing pattern}. To ease calculations in subsequent steps, the governing pattern is shifted in time to place the beginning of the curve in the time origin.
\subsubsection{Matching}
Matching process is meant to find the most similar part of the \emph{governing pattern}(called state) with the input data sequence. The last $H$ input data points are shifted in time within the time boundaries of the \emph{governing pattern} and the most similar state is chosen. To move the input sequence inside the time boundaries of the \emph{governing pattern}, the input sequence is shifted to the origin of time.
\begin{equation}\label{eqn:eq1}
	t_{i} \leftarrow  t_{i} - t_{n-H+1} ~~~, \; i = n-H+1, \ldots, n.
\end{equation}
In Eq. \eqref{eqn:eq1}, $t_{i}$ is the time variable of the $i^{th}$ input data point and $H$ is the length of the input stream and $t_{n}$ is the time variable of the last input data point. This step places the input stream in the origin of time.
\newline
Next, input data sequence will be shifted in time to find the most similar state.
\begin{equation}\label{eqn:eqsim}
	\begin{aligned}
	& \underset{t_{o}}{argmax} \sum_{i=n-H+1}^{n} similarity(X_{i,t_{o}},\phi_{pc}) \\
	& \text{subject to:} ~~~  P_{i,0} \leq t_{o} \leq P_{i,k};
	\end{aligned} 	
\end{equation}
In Eq. \eqref{eqn:eqsim}, the purpose is to find a $t_{o}$ as the beginning of the state in the \emph{governing pattern} having the maximum similarity with the data points in that state. $X_{i,t_{o}}=(t_{i}+t_{o},x_{i})$ and $(t_{i},x_{i})$ is the $i^{th}$ data point of the input sequence. Also, $t_{n}$ is the time upper bound of the input sequence and $\phi_{pc}$ represents the \emph{governing pattern}. $P_{i,0}$ and $P_{i,K}$ are the time variables of the first and the last points of the \emph{governing pattern}.

The method of computing similarity value determines the optimization method of the classifier. In the current paper, similarity value is computed through a modified KL-divergence. Eq. \eqref{eqn:eq3} implements the similarity function used in Eq. \eqref{eqn:eqsim}.
\begin{equation}\label{eqn:eq3}
	\begin{aligned}
	& \underset{t_{o}}{argmin} \sum_{i=n-H+1}^{n} \|F(X_{i,t_{o}})\| \log (\frac{\| F(X_{i,t_{o}}) - X_{i,t_{o}}\|}{\|F(X_{i,t_{o}})\|}) \\
	& ~~~~ \text{subject to:} ~~~  P_{i,0} \leq t_{o} \leq P_{i,k};
	\end{aligned} 
\end{equation}

where $P_{i,k}$ is the time value of last principal curve point and $P_{i,0}$ is the time value of first principal curve point. $X_{i,t_{o}}=(t_{i}+t_{o},x_{i})$ for each $t_{o}$.  $F(X_{i,t_{o}})$ is the projection of shifted data point on the nearest principal curve and $\tau$ is the iterator on the number of classes of data, which  is determined by $C$.
\newline
By defining potential areas, we are able to decrease the number of computations by searching potential timespans. We aim to find areas that the input stream is possible to have the highest similarity with the \emph{governing pattern}. The potential points are defined as Eq.\eqref{eqn:eqp} defines.
\begin{equation}\label{eqn:eqp}
	\begin{aligned}
		& M = E\{x_{n-H+1},...,x_{n}\}\\
	& \theta = \{P_{i} ~ |  ~   \| P_{i} - M \| / \|P_{i}\| \leq 0.2 \}\\
	& 
	\end{aligned} 	
\end{equation}
where $x_{n}$ is the last input data and $P_{i}$ is the $i^{th}$ point of the principal curve. $M$ is the expected value of the input data stream with $H$ data points, $\theta$ is the potential area for matching process.
\newline
By applying potential areas to the Eq. \eqref{eqn:eqsim}, the objective function changes as follows.
\begin{equation}\label{eqn:eqsimpot}
	\begin{aligned}
	& \underset{t_{o}}{argmax} \sum_{i=n-H+1}^{n} similarity(X_{i,t_{o}},\phi_{pc}) \\
	& \text{subject to:} ~~~  t_{o} \in \theta
	\end{aligned} 	
\end{equation}
Where $X_{i,t_{o}}=(t_{i}+t_{o},x_{i})$ and $(t_{i},x_{i})$ is the $i^{th}$ data point of the input sequence, $t_{n}$ is the time upper bound of the input sequence and $\phi_{pc}$ represents the \emph{governing pattern}. This process decreases computations with a considerable ratio.

\begin{figure*}[h]
\centering
\caption{\label{fig:matching}Example of matching process. This process finds a time shift which causes the most similarity between \emph{governing pattern} and the input sequence. Red colored points are the last H input data points and other colors define the curve of each class of data. Subfigure (a) is the desired time shift and the most similar situation where Subfigure (b) and Subfigure (c) are two wrong examples of matching process}
\subfloat[Correct Example]{%
  \includegraphics[width=0.32\textwidth,height=4cm]{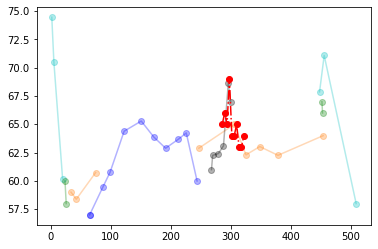}%
  \label{fig:match_0}%
}
\subfloat[Incorrect Example]{%
  \includegraphics[width=0.32\textwidth,height=4cm]{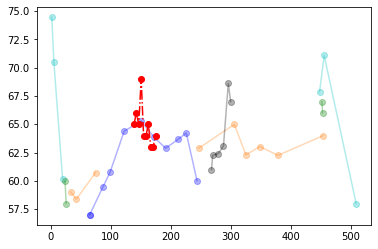}%
  \label{fig:match_1}%
}
\subfloat[Incorrect Example]{%
  \includegraphics[width=0.32\textwidth,height=4cm]{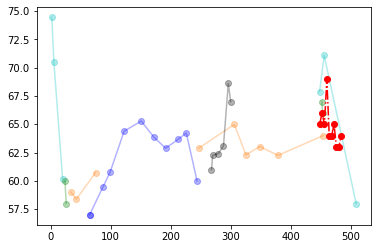}%
  \label{fig:macth_2}%
}
\end{figure*}
Fig. \ref{fig:matching} demonstrates matching process. In each figure, four principal curves shown in cyan, green, red and magenta colors represent the different classes of the data and blue sequence is the input data sequence. Horizontal axis is the time axis and vertical axis is one of the features of the initialization data. The correct answer demonstrated in the figure has the most similarity with the principal curves.
\par
As it is obvious, matching process makes no assumption about the time intervals between input data points and time intervals between initialization data points. This ability makes the proposed classifier able to recognize events when time span of the event changes.
\subsubsection{Label Prediction}
As stated before, one principal curve represents each class of the data. After Matching process and finding the most similar state, the last data point of the input sequence will be projected to all principal curves. The nearest principal curve will determine the predicted label. Eq. \eqref{eqn:eqlp} formulates this process.
\begin{equation}\label{eqn:eqlp}
 	\underset{\tau}{argmin} ~ \| f_{\tau}(X_{n,t_{o}})- X_{n,t_{o}} \|, ~~~\tau = 1,\ldots,C
\end{equation}
where $X_{n,t_{o}}$ is the last input data with optimized time shift and $f_{\tau}(X_{n,t_{o}})$ is the projection of $X_{n,t_{o}}$ on $\tau^{th}$ principal curve and $C$ is the number of classes.
\subsubsection{Online Learning}
Online learning helps classifier to modify the learned model and leads to better performance in prediction. Online learning is necessary when input data has intrinsic drift. The proposed algorithm draws a vector between the input data point(end of vector) and its projection on the nearest principal curve and moves the two neighboring principal curve points along drawn vector with a predefined step size($\alpha$). If predicted label is the same as the true label, the corresponding segment of the principal curve moves in the positive direction of the error vector. Eq. \eqref{eqn:eqol} demonstrates this process.
\begin{equation}\label{eqn:eqol}
	\begin{split}
	& k = \underset{i}{argmin}~~\{\| P_{i} -  f_{\tau}(X_{n,t_{o}}) \| ~ | ~ P_{i,0} > t_{n} \} \\
	& P_{k} ~~ +=  \alpha * (X_{n,t_{o}} - f_{\tau}(X_{n,t_{o}}))\\
	& P_{k-1} ~~ += \alpha * (X_{n,t_{o}} - f_{\tau}(X_{n,t_{o}}))\\
	\end{split}
\end{equation}
where $P_{i}$ is the $i^{th}$ point of principal curve and $P_{i,0}$ is the time variable of the corresponding curve point. Hence, $k$ is the first point of the principal curve which has a bigger time variable compared to $f_{\tau}(X_{n,t_{o}})$. $X_{n,t_{o}}$ is the input data point with optimized time shift and $\alpha$ is the step size of the learning process.
\newline
If the predicted label is different from the true label, the corresponding segment of the principal curve moves in the opposite direction of the error vector to decrease the probability of predicting the same wrong label in similar positions. Eq. \eqref{eqn:eqo2} demonstrates this process.
\begin{equation}\label{eqn:eqo2}
	\begin{split}
	& k = \underset{i}{argmin}~~\{\| p{i} -  f_{\tau}(X_{n,t_{o}}) \| ~ | ~ P_{i,0} > t_{n} \} \\
	& P_{k} ~~ -=  \alpha * (X_{n,t_{o}} - f_{\tau}(X_{n,t_{o}}))\\
	& P_{k-1} ~~ -= \alpha * (X_{n,t_{o}} - f_{\tau}(X_{n,t_{o}}))\\
	\end{split}
\end{equation}
Eq. \eqref{eqn:eqo2} has the same parameters as Eq. \eqref{eqn:eqo1} except that this process moves the curve points against the error vector.
\subsection{Selective sampling}
In addition to proposed steps, a deterministic selective sampling system can be built by defining a query strategy. After matching process, a similarity value will be produced which is the most similar state between input sequence and \emph{governing pattern}. By arrival of each data point, the produced similarity value will be added to a similarity history. Hence, by analyzing the variations in similarity values, system is able to decide which data is more worthy to query the label.
\subsubsection{Linear Query Strategy}
This strategy is a simple counting problem where the algorithm count the times that the similarity value has been decreased. By normalizing the value by the length of history, the produced value can be compared with a threshold
to decide on querying the label.
\begin{algorithm}[h]

\caption{Linear query strategy\label{alg:linearQ}}
    \SetAlgoLined
    \KwIn{$similarity^{1\times h}$}
    \Parameter{Threshold}
    \KwResult{queries label if return value is 1}
    $q = 1$\;
    \For{$i\leftarrow 2$ \KwTo $Length(similarity)$ }{
		\uIf{$ similarity[i] < similarity[i-1]$}{ 
	    	$q~ += 1$\;}
    }
    \uIf{$\dfrac{q}{Length(similarity)} \geq Threshold$}{ 
    		\Return 1\;}
    	\Else{
    		\Return 0\;}

\end{algorithm}
\newline
In Algorithm \ref{alg:linearQ}, parameter  $q$ is the return value of the algorithm and it will be initialized as 1. Starting from second value in similarity history, if current similarity value is less than the former value, $q$ will be incremented by one. This process will continue till the end of similarity history, then, $q$ will be divided by $Length(history)$ for normalization. Algorithm will query the label if $q > tr$ where $tr$ is the threshold for query process.
\subsubsection{Exponential Query Strategy}
In exponential strategy, the goal is to perform the counting with extra penalty for consecutive decrease in similarity values; such that after each decrease, penalty coefficient will be increase with ratio of 2 and increase in similarity value will decrease penalty coefficient with ratio of $\tfrac{1}{2}$. This strategy punishes consecutive decreases and rewards consecutive increases. Hence, in each step, penalty coefficient will be modified considering the similarity values and considering its value, system will be punished and at last, punishment value will be normalized and compared with a threshold to decide query the label.
\begin{algorithm}[t]
    \SetAlgoLined
    \KwIn{$similarity^{1\times h}$}
    \Parameter{Threshold}
    \KwResult{queries label if return value is 1}
    $c = 1$\;
    $q = 1$\;
    \For{$i\leftarrow 2$ \KwTo $Length(similarity)$ }{
    	$q *= c$\;
    	\uIf{$similarity[i] < similarity[i-1]$}{ 
    		$c \ =\ c*2$\;}
    	\Else{
    		$c \ =\ c/2$\;}
    }
    \uIf{$\dfrac{q}{2^{Length(similarity)-1}} \geq Threshold$}{ 
    		\Return 1\;}
    	\Else{
    		\Return 0\;}
\caption{Exponential query strategy \label{alg:expQ}}
\end{algorithm}
In Algorithm \ref{alg:expQ}, $q$ is the return value of query function and $c$ is penalty coefficient; both these values are initialized as $c,q = 1$. Starting from second element of the history, $q$ will be multiplied in $c$. If current similarity is less than the former one, $c$ will be multiplied in $2$, otherwise, in $\frac{1}{2}$ till the end of the similarity history. For normalizing $q$ value, it will be divided by $2^{Length(history)-1}$. Considering that $tr$ is the threshold for query process, algorithm will query the label if $q > tr$.

\section{Data description}
\subsection{WESAD dataset}
WESAD is a dataset for wearable stress and affect detection \cite{schmidt2018introducing}. This dataset contains samples from 15 patients each wearing one wrist-worn device, Empatica E4, and one chest-worn device, RespiBAN,. Here we just use data from the wrist-worn device to predict labels. The wrist-worn dataset contains the following features. Blood volume pulse (BVP) with 64 Hz sampling rate, electrodermal activity (EDA) with 4 Hz sampling rate, body temperature with 4 Hz sampling rate, and three-axis acceleration with 32 Hz sampling rate. Labels were also provided with 64 Hz frequency. 
\par
Given the different sampling rates, a dataset was constructed from the data corresponding to the least frequent points, which has no missing values. It is notable that no preprocessing was applied to the dataset. In terms of the occurrence of each label, this dataset is a balanced dataset.

\subsection{Sleep dataset}
The second dataset contains motion and heart rate features from an apple watch and was labeled according to sleep steps obtained from polysomnography data \cite{walch2019sleep}. Three acceleration axes and heart rate are the features that we want to use to predict the  sleep stage label in the dataset. Each subject wearing a device has spent one night in-lab for an 8-hour sleep and the labels were assigned using polysomnography. 5 stages naming wake, N1, N2, N3 and REM are demonstrated in dataset.
\newline
One approach for this dataset is the REM/Non-REM prediction which will be
highly imbalanced. Other approaches like Wake/Sleep have been tested on
this dataset, which are easier tasks. Because none of the features are linearly separable, for predicting REM/None-REM labels   analyzing the changes in features over time is the only way to predict labels. Patient 4018081 dataset is used for experiments.
\subsection{PPG dataset}
PPG-Dalia dataset is a multimodal dataset which features psychological and
motion data\cite{reiss2019deep}. The devices used in PPG-Dalia is the same as the devices used in WESAD dataset. As with WESAD, same features would be chosen to predict performed activities. Sitting, ascending or descending stairs, table soccer, cycling, driving a car, lunch break, walking and working are close to real-life actions which is performes by each subject. This dataset is assumed a balanced dataset. 
\newline
Compared to the previous datasets, PPG has a sensible drift in the data, which will be helpful in testing the online learning process of the proposed algorithm. The drift in data is duo to the fact that two data batches from two different patients with similar parameters are used as our dataset for experiments.
\subsection{PAMAP dataset}
PAMAP dataset is a physical activity monitoring dataset \cite{reiss2012introducing}. This dataset is sampled from 9 subjects via wearable datasets. each subject wears three inertial measurement units (IMU) and a heart rate sensor. subjects have some predefined activities to do in sequence, which are lying, sitting, standing, walking, running, cycling, Nordic walking, watching TV, computer work, car driving, ascending stairs, descending stairs, vacuum cleaning, ironing, folding laundry, house cleaning, playing soccer, rope, jumping and transient activities.
Each data point has 52 features, a heart rate and 17 features for each IMU sensor. Taking into account the number of occurrences per label, this dataset is considered balanced.
\section{experimental results}
\subsection{Classifier}
In this section, the proposed algorithm has been compared with three state-of-the-art algorithms for time-series prediction. Implementation details of these three neural networks are reported in Table \ref{tab:nn}. In addition to accuracy, considering imbalanced datasets, F-score and G-score have been used for evaluating classifiers. F-score is calculated as follows.
\begin{equation}\label{eqn:fsc}
 	\begin{split}
	& F_{\beta}^{\ell}=\frac{precision^{\ell}.recall^{\ell}}{\beta^{2}.precision^{\ell}+recall^{\ell}} \\
	& =\frac{(1+\beta^{2}).TP^{\ell}}{(1+\beta^{2}).TP^{\ell}+FP^{\ell}+\beta^{2}.FN^{\ell}}\\
	\end{split}
\end{equation}
\begin{equation}
	F_{\beta}=\frac{1}{N} \sum_{\ell=1,...,N} C^{\ell}.F_{\beta}^{\ell}
\end{equation}
$F^{\ell}_{\beta}$ is the F-score of class $\ell$ with hyper parameter $\beta$ and 
$C^{\ell}$ is the importance of class $\ell$ while N is the number of classes. $TP^{\ell}$, $FP^{\ell}$ and $FN^{\ell}$ represent true positive, false positive and false negative numbers of the class $\ell$. In experiments, $C_{\ell} = 1$ and $\beta = 1$.
\newline
G-score is the generalization of Jaccard measure and is calculated as follows.
\begin{equation}\label{eqn:gsc}
 	G_{\beta}^{\ell}=\frac{TP^{\ell}}{TP^{\ell}+FP^{\ell}+\beta.FN^{\ell}}
\end{equation}
\begin{equation}
	G_{\beta}=\frac{1}{N} \sum_{\ell=1,...,N} C^{\ell}.G_{\beta}^{\ell}
\end{equation}
$G^{\ell}_{\beta}$ is the G-score of class $\ell$ with hyper parameter $\beta$ and 
$C_{\ell}$ is the importance of class $\ell$ while N is the number of classes. In experiments, $C_{\ell} = 1$ and $\beta = 1$.
\newline
As demonstrated in Eq. \eqref{eqn:fsc} and Eq. \eqref{eqn:gsc}, G-score penalizes false predicted instances more than F-score. In other words, for imbalanced data, F-score reports the average performance while G-score reports the worst-case performance. Hence, both metrics have been reported in result tables.

\begin{table*}[t]
\begin{center}
	\captionof{table}{\label{tab:nn} details of implemented neural network architectures used for comparison in experimental results}
	\begin{tabularx}{\textwidth}{ c  L } 
	\toprule
	Neural network & Description\\
	\midrule
	LSTM  & 80 LSTM cells with a fully connected layer on top of LSTM network
Network is trained 100 epochs on initialization data and 40 epochs on each 10 data points batch for online learning. Optimizer is Adam and loss function is MAE. History length is set to 10.\\
	\midrule
	GRU& 80 GRU cells form recurrent structure. A fully connected layer connects GRU outputs to output. Network is trained 100 epochs on initialization data and 40 epochs on each batch of data during online learning. Optimizer is Adam and loss function is MAE. History length is set to 10. \\
	\midrule
	TCN & A TCN network with 64 filters and dilation of powers of 2. Network is trained 100 epochs on initialization data and 40 on each batch of online learning data. A fully connected layer connects TCN output to one-hot-encoded output.  Optimizer is Adam and loss function is MAE. History length is set to 10.\\
	\bottomrule
	\end{tabularx}
	
\end{center}
\end{table*}
\par
Sampling rate is the parameter to control the time span of an event. \textbf{Changing sampling rate allows us to simulate time dilation phenomenon}. Simply put, $sampling rate = \frac{1}{c}$ means that we only sample first data point out of each $c$ consecutive data points. For instance, if $c$ is chosen greater than $1$, the sampled event has same time span with less data points which their time difference between each two data points has been multiplied by $c$. In other words, we intend to test classifiers when data points have been reduced with ratio of $c$.

\begin{subsubsection}{WESAD dataset}
First experiment is on the WESAD dataset. $60\%$ of data points are used as training data and the rest is used as test data. History length is set to 10 for all classifiers and $sampling rate=1$ in current experiment. All tables report time in seconds.
\begin{center}
	\captionof{table}{\label{tab:wes1} WESAD results with sampling rate = 1}
	\begin{tabular}{ c  c  c  c  c} 
	\toprule
	&Accuracy & F-score & G-score & Time\\
	\midrule
	Proposed Alg.& \textbf{1.0} & \textbf{1.0} & \textbf{1.0} & 2434\\
	LSTM& 0.9926 & 0.9945 & 0.9892 & 1853\\
	GRU & 0.9460 & 0.9563 & 0.9181 & 1663\\
	TCN & 0.8743 & 0.8635 & 0.7626 & \textbf{1403} \\
	\bottomrule
	\end{tabular}
	 
\end{center}

\par
In next experiment, by choosing $sampling rate=\frac{1}{10}$, we aim to test classifiers under reduction of data points. For feeding reduced dataset into the neural networks, we used two settings. In $1^{st}$ setting, reduced dataset will be fed with no modification and in $2^{nd}$ setting, a linear function between each two data points approximates value of omitted data points and at last number of data points will be identical to the original (not reduced) dataset. History length will be 10 for classifiers.

\begin{center}
	\captionof{table}{\label{tab:wes2} WESAD results with sampling rate = $\frac{1}{10}$}
	\begin{tabular}{ c  c  c  c  c  }
	\toprule
	& Accuracy & f-score & G-score & Time \\
	\midrule
	Proposed Alg. & \textbf{1.0} & \textbf{1.0} & \textbf{1.0} & 2389\\
	LSTM-$1^{st}setting$  & 0.9419 & 0.9248 & 0.8629 & 1946 \\
	LSTM-$2^{nd}setting$  & 0.9886 & 0.9866 & 0.9736 & 1958 \\
	GRU-$1^{st}setting$ & 0.9451 & 0.9331 & 0.8774 & 1745 \\
	GRU-$2^{nd}setting$ & 0.9803 & 0.9762 & 0.9539 &  1803 \\
	TCN & 0.8258 & 0.8185 & 0.7010 & \textbf{1493} \\
	\toprule
	\end{tabular}
	
\end{center}

By setting $sampling rate=\frac{1}{30}$results in the following table.
\begin{center}
	\captionof{table}{\label{tab:wes3} WESAD results with sampling rate = $\frac{1}{30}$}
	\begin{tabular}{ c  c  c  c  c  }
	\toprule
	& Accuracy & f-score & G-score & Time \\
	\midrule
	Proposed Alg. & \textbf{0.9969} & \textbf{0.9976} & \textbf{0.9952} & 2524 \\
	LSTM-$1^{st}setting$  & 0.7822 & 0.6194 & 0.5081 & 2766\\
	LSTM-$2^{nd}setting$  & 0.9850 & 0.9703 & 0.9438 & 2815 \\
	GRU-$1^{st}setting$ & 0.8189 & 0.8178 & 0.6923 & \textbf{2174} \\
	GRU-$2^{nd}setting$ & 0.9799 & 0.9793 & 0.9539 & 2364 \\
	TCN & 0.5592 & 0.5562 & 0.3924 & 2218\\
	\toprule
	\end{tabular}
	
\end{center}
In last experiment on WESAD dataset, intervals between each two samples is determined by a random number. In other words, $sampling rate$ changes randomly in each data point selection. Here, $sampling rate$ is chosen by random sampling from integer number between 1 and 30. Other parameters of the classifiers are the same.
\begin{center}
	\captionof{table}{\label{tab:wes4} WESAD results with random sampling rate between 1 and $\frac{1}{30}$}
	\begin{tabular}{ c  c  c  c  c }
	\toprule
	& Accuracy & f-score & G-score & Time \\
	\midrule
	Proposed Alg. & \textbf{0.9973} & \textbf{0.9977} & \textbf{0.9958} & 2328 \\
	LSTM-$1^{st}setting$  & 0.6436 & 0.5211 & 0.4039 & 2362\\
	LSTM-$2^{nd}setting$  & 0.9803 & 0.9714 & 0.9697 & 2395 \\
	GRU-$1^{st}setting$ & 0.8327 & 0.7983 & 0.6692 & 2116 \\
	GRU-$2^{nd}setting$ & 0.9588 & 0.9537 & 0.9505 & 2294\\
	TCN & 0.6337 & 0.5935 & 0.4631 & \textbf{1949}\\
	\toprule
	\end{tabular}
	
\end{center}
\par
As shown in Tab. \ref{tab:wes1}, Tab.\ref{tab:wes2} and Tab.\ref{tab:wes3}, changing sampling rates causes more wrong predictions in the classifier even if the skipped data points were approximated with a linear equation. Whereas the proposed algorithm is slightly affected by changes in the sampling rate. However results show that the proposed algorithm is more time consuming compared to other algorithms.
\end{subsubsection}
\begin{subsubsection}{Sleep dataset}
The second dataset is sleep stage prediction dataset which is highly imbalanced. $60\%$ of data points is used as training data and the rest is used as test data. History length is 10 and $sampling rate=1$. Table \ref{tab:sleep1} reports the results.

\begin{center}
	\captionof{table}{\label{tab:sleep1} Sleep results with sampling rate = 1}
	\begin{tabular}{ c  c  c  c  c}
	\toprule
	& Accuracy & f-score & G-score & Time \\
	\midrule
	Proposed Alg. & \textbf{0.9707} & \textbf{0.8423} & \textbf{0.7289} & 3218 \\
	LSTM & 0.8957 & 0.6883 & 0.5717 & 1694\\
	GRU & 0.9358 & 0.7851 & 0.6488 & 1531 \\
	TCN & 0.8520 & 0.2105 & 0.1176 & \textbf{1204}\\
	\toprule
	\end{tabular}
	
\end{center}
By changing sampling rate to $\frac{1}{2}$, results will be as follows.
\begin{center}
	\captionof{table}{\label{tab:sleep2} Sleep results with sampling rate = $\frac{1}{2}$}
	\begin{tabular}{ c  c  c  c  c}
	\toprule
	& Accuracy & f-score & G-score & Time \\
	\midrule
	Proposed Alg. & 0.9680 & \textbf{0.8849} & \textbf{0.7935} & 3315 \\
	LSTM-$1^{st}setting$  & 0.8876 & 0.1428 & 0.0769 & 1189\\
	LSTM-$2^{nd}setting$  & 0.9561 & 0.6551 & 0.4871 & 1326  \\
	GRU-$1^{st}setting$ & \textbf{0.9785} & 0.7104 & 0.5508 & 1173 \\
	GRU-$2^{nd}setting$ & 0.9618 & 0.7642 & 0.6184 & 1262\\
	TCN & 0.7280 & 0.1987 & 0.1103 & \textbf{1081}\\
	\toprule
	\end{tabular}
	
\end{center}

Results with the sampling rate=$\frac{1}{4}$ is as follows.
\begin{center}
	\captionof{table}{\label{tab:sleep3} Sleep results with sampling rate = $\frac{1}{4}$}
	\begin{tabular}{ c  c  c  c  c}
	\toprule
	& Accuracy & f-score & G-score & Time \\
	\midrule
	Proposed Alg. & 0.9778 & \textbf{0.8686} & \textbf{0.7678} & 3104 \\
	LSTM-$1^{st}setting$  & 0.8913 & 0.0153 & 0.0077 & 1719\\
	LSTM-$2^{nd}setting$  & \textbf{0.9802} & 0.4960 & 0.3298 & 1761  \\
	GRU-$1^{st}setting$ & 0.9770 & 0.2189 & 0.1229 & 1549 \\
	GRU-$2^{nd}setting$ & 0.9666 & 0.7213 & 0.5641 & 1691\\
	TCN & 0.3197 & 0.0403 & 0.0206 & \textbf{1381}\\
	\toprule
	\end{tabular}
		
\end{center}
In last experiment, in each step, a random number between $1$ and $4$ determines the sampling rate. The results are demonstrated in Table \ref{tab:sleep4}.
\begin{center}
	\captionof{table}{\label{tab:sleep4} Sleep results with random sampling rate between 1 and $\frac{1}{4}$}
	\begin{tabular}{ c  c  c  c  c}
	\toprule
	& Accuracy & f-score & G-score & Time \\
	\midrule
	Proposed Alg. & \textbf{0.9855} & \textbf{0.8926} & \textbf{0.8061} & 2881 \\
	LSTM-$1^{st}setting$  & 0.9008 & 0.0176 & 0.0089 & 1541\\
	LSTM-$2^{nd}setting$  & 0.9261 & 0.4917 & 0.3489 & 1533  \\
	GRU-$1^{st}setting$ & 0.9127 & 0.1184 & 0.0629 & 1472 \\
	GRU-$2^{nd}setting$ & 0.8693 & 0.7152 & 0.5601 & 1362\\
	TCN & 0.9180 & 0.2130 & 0.1192 & \textbf{1194}\\
	\toprule
	\end{tabular}
		
\end{center}
Results of sleep stage prediction dataset confirm that the proposed algorithm can handle any variation in the time difference of the input data points. In other words, the relative concept of time is involved in label prediction process of the proposed algorithm. The significant difference between results of the proposed algorithm and other neural networks is because that sleep stage prediction dataset is highly imbalanced and the proposed algorithm does not loose accuracy on imbalanced data. The proposed algorithm is immune to imbalanced data because that no matter how many of the data points belong to the minority class, the occupied time span of the minority data points will not be violated by the majority class. As shown in results, in some situations, neural network classifiers have higher accuracy but lower F-score and G-score which confirms the stated assertion about the imbalanced data.
\end{subsubsection}
\begin{subsubsection}{PPG dataset}
\par
The third experiment is on PPG dataset. $60\%$ of data points is used as training data and the rest is used as test data. History length is 10 and $sampling rate=1$. Table \ref{tab:ppg1} reports the results.
\begin{center}
	\captionof{table}{\label{tab:ppg1} PPG results with sampling rate = 1}
	\begin{tabular}{ c  c  c  c  c}
	\toprule
	& Accuracy & f-score & G-score & Time \\
	\midrule
	Proposed Alg. & 0.7737 & 0.5896 & 0.4904 & 5306 \\
	LSTM  & 0.8078 & 0.6175 & 0.5152 & 2178\\
	GRU & \textbf{0.8296} & 0.6172 & 0.5255 & 1994 \\
	TCN & 0.7482 & \textbf{0.7287} & \textbf{0.5784} & \textbf{1762}\\
	\toprule
	\end{tabular}
		
\end{center}

In following table, the $sampling rate=\frac{1}{60}$.
\begin{center}
	\captionof{table}{\label{tab:ppg2} PPG results with sampling rate = $\frac{1}{60}$}
	\begin{tabular}{ c  c  c  c  c}
	\toprule
	& Accuracy & f-score & G-score & Time \\
	\midrule
	Proposed Alg. & \textbf{0.8787} & \textbf{0.5975} & \textbf{0.5183} & 4726 \\
	LSTM-$1^{st}setting$  & 0.6329 & 0.3458 & 0.2563 & 2759\\
	LSTM-$2^{nd}setting$  & 0.8036 & 0.5812 & 0.4754 & 2853  \\
	GRU-$1^{st}setting$ & 0.6347 & 0.3190 & 0.2337 & 2461 \\
	GRU-$2^{nd}setting$ & 0.6458 & 0.4190 & 0.3037 & 2614\\
	TCN & 0.5448 & 0.2709 & 0.1855 & \textbf{2041}\\
	\toprule
	\end{tabular}
		
\end{center}

In following table, the $sampling rate=\frac{1}{90}$.
\begin{center}
	\captionof{table}{\label{tab:ppg3} PPG results with sampling rate = $\frac{1}{90}$}
	\begin{tabular}{ c  c  c  c  c}
	\toprule
	& Accuracy & f-score & G-score & Time \\
	\midrule
	Proposed Alg. & \textbf{0.8627} & \textbf{0.5710} & \textbf{0.4841} & 4319 \\
	LSTM-$1^{st}setting$  & 0.4506 & 0.3045 & 0.1927 & 3481\\
	LSTM-$2^{nd}setting$  & 0.6960 & 0.4123 & 0.3005 & 3589  \\
	GRU-$1^{st}setting$ & 0.4822 & 0.3098 & 0.2073 & 3183 \\
	GRU-$2^{nd}setting$ & 0.6092 & 0.3788 & 0.2619 & 3503\\
	TCN & 0.3645 & 0.1897 & 0.1168 & \textbf{2918}\\
	\toprule
	\end{tabular}
	
\end{center}
And in last experiment on PPG, in each time step, $sampling rate$ is an integer random number between 1 and 90.
\begin{center}
	\captionof{table}{\label{tab:ppg4} PPG results with random sampling rate between 1 and $\frac{1}{90}$}
	\begin{tabular}{ c  c  c  c  c}
	\toprule
	& Accuracy & f-score & G-score & Time \\
	\midrule
	Proposed Alg. & \textbf{0.8692} & \textbf{0.5803} & \textbf{0.5060} & 4518 \\
	LSTM-$1^{st}setting$  & 0.4891 & 0.3014 & 0.1895 & 2798\\
	LSTM-$2^{nd}setting$  & 0.7315 & 0.4672 & 0.3599 & 2674 \\
	GRU-$1^{st}setting$ & 0.4340 & 0.2936 & 0.1964 & 2418 \\
	GRU-$2^{nd}setting$ & 0.6782 & 0.3979 & 0.3145 & 2547\\
	TCN & 0.3918 & 0.2624 & 0.2173 & \textbf{2381}\\
	\toprule
	\end{tabular}
	
\end{center}

\par
Results of PPG datset confirms the former conclusions on WESAD and sleep datasets. The unexpected results that have occurred in this experiment is that with decreasing the sampling rate, all evaluation parameters of the proposed algorithm have been increased; Tab. \ref{tab:ppg1} and Tab. \ref{tab:ppg2} demonstrate this phenomenon. By reducing sampling rate and keeping the history length the same, the input stream covers a wider time span. In dense datasets, like PPG, wider time span results in better matching process and hence better label prediction. This eider time span justifies the peculiar results of the PPG dataset. However, there should be a trade-off between these two parameters.
\newline
In Tab. \ref{tab:ppg1}, TCN ,GRU and LSTM have better performance compared to the proposed algorithm. Considering that the PPG dataset has intrinsic drift, which former dataset did not have, it is possible that the proposed algorithm is not capable of handling drift in data stream. In other words, online learning step of the proposed algorithm is not functioning as expected.

\end{subsubsection}

\subsection{Analysis of Online Learning}
In this experiment, for analyzing online learning step of the proposed algorithm, we aim to compare classifiers without learning from the test data. In other words, classifier only learns from initialization data points. Sampling rate is equal to 1 in this experiment and other parameters are same as the previous experiments.

\begin{table}[h]
\begin{center}
	\captionof{table}{\label{tab:senza-online1} Results of the proposed classifier without online learning on test data. sampling rate is set to 1 for this experiment.}
	\begin{tabularx}{0.43\textwidth}{ c | c | c | c | c | c } 
	\toprule
	Dataset & & Accuracy & f-score & G-score & Time \\\hline
	\multirow{4}{*}{WESAD} & TPM& \textbf{0.9842} & \textbf{0.9639} & \textbf{0.9619} & \textbf{2016}\\
						    & LSTM& 0.9661 & 0.9570 & 0.9505 & 1628\\				    
						    & GRU & 0.9179 & 0.8992 & 0.8879 & 1408\\
						    & TCN & 0.8287 & 0.8149 & 0.7238 & 1284 \\
                            \hline
    \multirow{4}{*}{Sleep} & TPM & \textbf{0.9625} & \textbf{0.7945} & \textbf{0.7018} & \textbf{2817} \\
						    & LSTM & 0.8186 & 0.3972 & 0.2610 & 1192\\
						    & GRU & 0.9103 & 0.4913 & 0.3172 & 1092 \\
						    & TCN & 0.8017 & 0.1983 & 0.1147 & 1028\\
                            \hline
	\multirow{4}{*}{PPG} & TPM & \textbf{0.7162} & \textbf{0.5162} & \textbf{0.4261} & \textbf{4719} \\
						    & LSTM  & 0.7028 & 0.4961 & 0.4285 & 1863\\				    
						    & GRU & 0.7168 & 0.5219 & 0.4381 & 1639 \\
						    & TCN & 0.6838 & 0.6139 & 0.5261 & 1529\\

	\bottomrule
	\end{tabularx}
	
\end{center}
\end{table}

Comparing results of Tab.\ref{tab:senza-online1} with Tab. \ref{tab:wes1}, Tab.\ref{tab:sleep1} and Tab.\ref{tab:ppg1} shows that in all situation, the proposed classifier has closer accuracy, F-score and G-score with the situation that classifiers are rained with the test data in online learning. Highlighted values in Tab.\ref{tab:senza-online1} show the least difference between current setting and the experiments of part A. In other words, TPM is more vulnerable to drifts in data. These results confirms the results of Tab. \ref{tab:ppg1} where on sampling rate 1, the proposed classifier had weaker performance than neural networks; because PPG dataset has sensible drift in data compared to WESAD and Sleep datasets.
\par
To evaluate the effect of step size, $\alpha$, in online learning, in following experiment, $\alpha$ varies from $[0.2,2]$ by footstep equal to 0.1. Tab. \ref{tab:alpha1} reports best and mean accuracy of the proposed classifier while alpha is changing in $[0.2,2]$.

\begin{table}[h]
\begin{center}
	\captionof{table}{\label{tab:alpha1} Results of the proposed classifier while $\alpha$ parameter in online learning changes in range $[0.2,2]$ with footstep $0.1$. Values are extracted from 10 runs of the algorithm.}
	\begin{tabularx}{0.48\textwidth}{ c | c c | c c | c c } 
	\toprule
	 & \multicolumn{2}{c}{Accuracy} & \multicolumn{2}{c}{f-score} & \multicolumn{2}{c}{G-score} \\\hline
	Dataset & best & mean & best & mean & best & mean\\\hline
	WESAD & 1.0 & 0.9832 & 1.0 & 0.9513 & 1.0 & 0.9037\\

    Sleep & 0.9810 & 0.9263 & 0.8493 & 0.7318 & 0.7361 & 0.6218\\

	PPG  & 0.7759 & 0.6818 & 0.5898 & 0.4962 & 0.4912 & 0.4904 \\

	\bottomrule
	\end{tabularx}
	
\end{center}
\end{table}

As shown in Tab.\ref{tab:alpha1}, considering the difference between mean value and best value of the metrics, online learning in the proposed classifier is highly dependent to $\alpha$ parameter. Nevertheless, online learning step in the proposed algorithm has not the expected functionality considering its time consumption.

\subsection{Selective sampling}
Following experiment will compare the proposed algorithm for selective sampling with winnow selective sampling and OA3L algorithm. The classifier used for all sampling algorithms in following experiment is the proposed classifier. All experiments will simulate an online sampler device where there exists an initialization data will all corresponding labels and unlabeled test data points which will be fed to the classifier in an online manner. In arrival of each data point, classifier predicts the label and query strategy decides the query of the corresponding label.
\par
Each reported value in result tables is \textbf{mean value $\pm$ variance} of data from 10 runs of each query strategy. Table \ref{tab:metrics} explains the metrics used in tables.
\begin{table}[h]
\begin{center}
	\captionof{table}{\label{tab:metrics} used metrics}
	\begin{tabularx}{0.5\textwidth}{ c  L } 
	\toprule
	metrics & Description\\
	\midrule
	apt query  & Queried labels that predicted label is different from true label\\
	\midrule
	inapt query& Queried labels that predicted label is equal to true label \\
	\midrule
	ratio & ratio of good queries to total queries $\dfrac{aptQuery}{aptQuery + inaptQuery}$\\
	\bottomrule
	\end{tabularx}
\end{center}
\end{table}
\newline
Implementation details are as follows. For winnow algorithm, $\eta = 2$ and $b = 2$. In OA3L algorithm, $\delta_{+}=0.2$ and $\delta_{-}=0.2$ for PPG, PAMAP and WESAD datasets; and considering that sleep dataset is imbalanced, $\delta_{+}=0.2$ and $\delta_{-}=0.4$.

\subsubsection{Quality of queried labels}
In first experiment, the proposed algorithm with linear and exponential strategies will be compared with OA3L and winnow algorithms. $10\%$  of data points are used as initialization data, where all corresponding labels are provided, and $90\%$ of data is used as test data where no labels are provided and there exists a budget equal to 30 queries for each 100 input data points. Table \ref{tab:results4} contains the results.

\renewcommand{\arraystretch}{1.2}
\begin{table*}[ht]
\begin{center}
	\captionof{table}{\label{tab:results4} Results on quality of query attempts with budget 30}
	\begin{tabularx}{0.80\textwidth}{ c | c | c | c | c | c } 
	\toprule
	 & Dataset & PPG & WESAD & Sleep & PAMAP \\\hline
	\multirow{3}{*}{Proposed algorithm Linear} & apt query & 5.46 $\pm$ 0.57 & 5.92 $\pm$ 0.73 &7.79 $\pm$ 1.05 & 7.91 $\pm$ 0.49\\
                                 & inapt query & 13.84 $\pm$ 1.14 & 18.81 $\pm$ 1.97 & 16.78 $\pm$ 1.54 & 15.05 $\pm$ 1.61\\
                                 & ratio & 0.28 $\pm$ 0.03 & 0.23 $\pm$ 0.07 & 0.32 $\pm$ 0.04 & \textbf{0.34 $\pm$ 0.02}\\\hline
    \multirow{3}{*}{Proposed algorithm Exponential} & apt query & 5.80 $\pm$ 0.75 & 6.23 $\pm$ 0.64 & 7.14 $\pm$ 0.58 & 6.01 $\pm$ 0.54\\
                                 & inapt query & 14.20 $\pm$ 1.32 & 18.65 $\pm$ 2.10 & 14.57 $\pm$ 1.17 & 16.28 $\pm$ 2.21\\
                                 & ratio & \textbf{0.29 $\pm$ 0.04} & \textbf{0.25 $\pm$ 0.07} & \textbf{0.33 $\pm$ 0.03} & 0.27 $\pm$ 0.05\\\hline
	\multirow{3}{*}{winnow algorithm} & apt query & 4.14 $\pm$ 1.64 & 3.46 $\pm$ 1.96 &7.46 $\pm$ 2.64 &6.14 $\pm$ 2.46\\
                                 & inapt query & 13.28 $\pm$ 1.83 & 16.28 $\pm$ 2.98 &16.18 $\pm$ 2.63 &13.94 $\pm$ 1.92\\
                                 & ratio & 0.23 $\pm$ 0.07 & 0.17 $\pm$ 0.08 & 0.32 $\pm$ 0.07 & 0.31 $\pm$ 0.09\\\hline
	\multirow{3}{*}{OA3L algorithm} & apt query & 6.8 $\pm$ 4.30 & 5.70 $\pm$ 4.58 & 4.80 $\pm$ 2.45 & 6.58 $\pm$ 2.83\\
                                 & inapt query & 18.40 $\pm$ 3.44 & 19.30 $\pm$ 4.10 &17.20 $\pm$ 2.82 &17.79 $\pm$ 3.06\\
                                 & ratio & 0.26 $\pm$ 0.15 & 0.23 $\pm$ 0.17 & 0.26 $\pm$ 0.22 & 0.27 $\pm$ 0.10\\
		
	\bottomrule
	\end{tabularx}
	
\end{center}
\end{table*}

In second experiment, the budget for query is equal to 90 for every 1000 input data points. Table \ref{tab:results5} contains the results.

\begin{table*}[h]
\begin{center}
	\captionof{table}{\label{tab:results5} Results on quality of query attempts with budget 90}
	\begin{tabularx}{0.80\textwidth}{ c | c | c | c | c | c } 
	\toprule
	 & Dataset & PPG & WESAD & Sleep & PAMAP \\\hline
	\multirow{3}{*}{Proposed algorithm Linear} & apt query & 17.41 $\pm$ 1.01 & 23.01 $\pm$ 1.14 & 27.35 $\pm$ 2.67 & 24.31 $\pm$ 2.07\\
                                 & inapt query  & 32.29 $\pm$ 1.36 & 34.85 $\pm$ 1.92& 49.73 $\pm$ 2.83 & 49.12 $\pm$ 2.79\\
                                 & ratio  &0.35 $\pm$ 0.06& 0.39 $\pm$ 0.06& 0.35 $\pm$ 0.07 &0.33 $\pm$ 0.06\\\hline
    \multirow{3}{*}{Proposed algorithm Exponential} & apt query & 19.41 $\pm$ 1.05& 23.73 $\pm$ 1.45& 26.65 $\pm$ 2.07& 27.35 $\pm$ 1.72\\
                                 & inapt query &30.52 $\pm$ 1.82& 32.05 $\pm$ 1.76& 47.03 $\pm$ 2.74& 48.28 $\pm$ 3.12\\
                                 & ratio & \textbf{0.38 $\pm$ 0.07}& \textbf{0.42 $\pm$ 0.03}& \textbf{0.36 $\pm$ 0.05}& \textbf{0.36 $\pm$ 0.06}\\\hline
	\multirow{3}{*}{winnow algorithm} & apt query & 16.03 $\pm$ 2.30& 20.70 $\pm$ 2.24 &26.15 $\pm$ 2.73 &29.15 $\pm$ 2.43\\
                                 & inapt query & 37.50 $\pm$ 2.83 &34.05 $\pm$ 3.06 &56.83 $\pm$ 3.41 &51.83 $\pm$ 3.01\\
                                 & ratio &  0.29 $\pm$ 0.07 & 0.38 $\pm$ 0.07 & 0.32 $\pm$ 0.09 &\textbf{0.36 $\pm$ 0.08}\\\hline
	\multirow{3}{*}{OA3L algorithm} & apt query & 16.42 $\pm$ 6.28 & 27.92 $\pm$ 7.28& 22.08 $\pm$ 4.70& 18.15 $\pm$ 4.28\\
                                 & inapt query & 39.80 $\pm$ 5.63 &54.06 $\pm$ 8.40 &59.41 $\pm$ 5.16 &58.37 $\pm$ 4.71\\
                                 & ratio & 0.29 $\pm$ 0.15 & 0.34 $\pm$ 0.19 & 0.25 $\pm$ 0.17 & 0.23 $\pm$ 0.16\\
		
	\bottomrule
	\end{tabularx}
	
\end{center}
\end{table*}
\par
\begin{figure}[h]
	\caption{\label{fig:res_0} Mean ratio of query quality with budget 30}
    \center{\includegraphics[width=\linewidth]{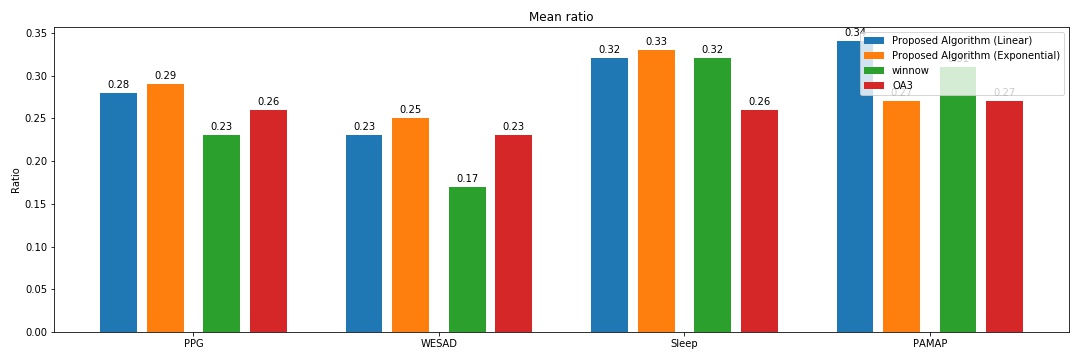}}    
\end{figure}
\begin{figure}[h]
	\caption{\label{fig:res_1} Variance ratio of query quality with budget 30}
    \center{\includegraphics[width=\linewidth]{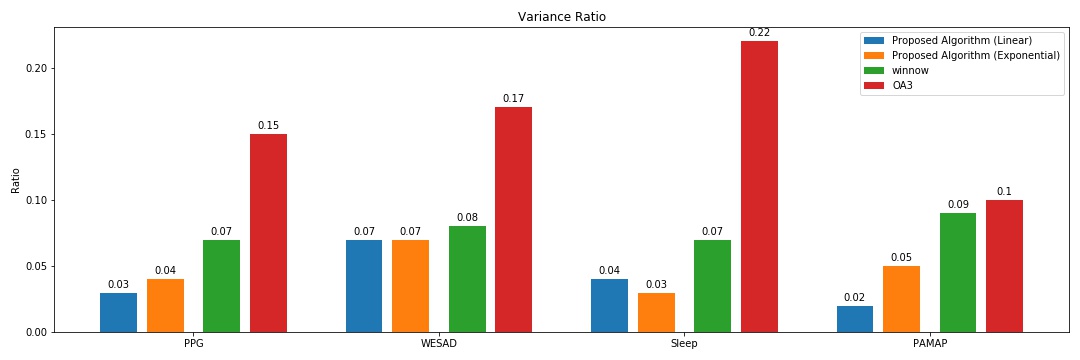}}    
\end{figure}
\begin{figure}[h]
	\caption{\label{fig:res_2} Mean and Variance ratio of query quality with budget 30}
    \center{\includegraphics[width=\linewidth]{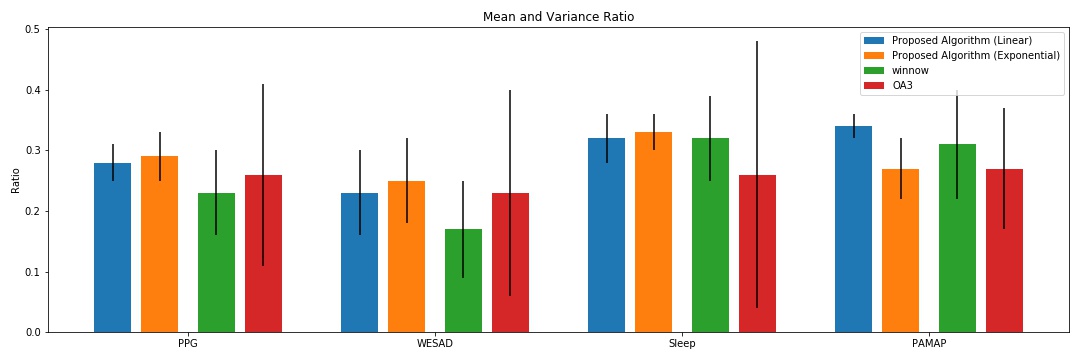}}    
\end{figure}

Results of Tab. \ref{tab:results4} and Tab. \ref{tab:results5} show that the proposed algorithm has more query requests in situations that the predicted label is not the same true label, which we call an apt query. Also, results show that the proposed algorithms have less variance in all situations compared to two other sampling algorithms. This low variance of the proposed algorithm is caused by removing the randomness from the query strategy which accelerates reliability of the sampling strategy in intensive risk environments. Even in situations that other algorithms had better ratio than one of the proposed algorithms, the lower variance belongs to the proposed algorithm. Results of Tab.\ref{tab:results4} are summarized in \ref{fig:res_0},\ref{fig:res_1} and \ref{fig:res_2}.

\subsubsection{Accuracy tests}
In these experiments, effect of different query algorithms on accuracy of the classifier will be evaluated. In first experiment, $50\%$ of data points with corresponding labels are used as initialization data and the rest of data points are used as test data with query budget equal to 90 for each 1000 data points. Initialization accuracy reports the accuracy of the classifier on initialization data and test accuracy reports the accuracy of classifier on test data while query budget is limited. Table \ref{tab:acc1} reports the results of this experiment.

\begin{table*}[h]
\begin{center}
	\captionof{table}{\label{tab:acc1} Accuracy results on Initialization data with full budget and test data with limited budget}
	\begin{tabularx}{0.80\textwidth}{ c | c | c | c | c | c } 
	\toprule
	 & Dataset & PPG & WESAD & Sleep & PAMAP \\\hline
	\multirow{3}{*}{Proposed algorithm Linear} & Init. Acc. & 63.50 $\pm$ 1.76 & 93.36 $\pm$ 1.87 & 73.30 $\pm$ 2.01 & 78.25 $\pm$ 1.97\\
							                 & Test Acc.  & 67.24 $\pm$ 1.83 & 94.16 $\pm$ 1.36& 74.86 $\pm$ 1.88 & 79.98 $\pm$ 2.53\\
                            				 & progress  & + 3.75 & +0.80 & \textbf{+1.56} &+1.73 \\\hline
    \multirow{3}{*}{Proposed algorithm Exponential} & Init. Acc. & 64.52 $\pm$ 1.93& 91.64 $\pm$ 1.99& 71.78 $\pm$ 2.16& 76.47 $\pm$ 1.35\\
							                     & Test Acc. & 68.47 $\pm$ 1.39& 93.15 $\pm$ 1.89& 72.95 $\pm$ 2.21& 79.83 $\pm$ 2.61\\
                            				     & progress & \textbf{+ 3.95 }& \textbf{+1.51 }& +1.17 & \textbf{+3.36 }\\\hline
	\multirow{3}{*}{winnow algorithm} & Init. Acc. & 64.2 $\pm$ 2.48& 95.22 $\pm$ 2.88 &74.20 $\pm$ 3.13 &77.85 $\pm$ 2.93\\
	                                 & Test Acc. & 62.65 $\pm$ 2.06 &93.29 $\pm$ 3.31 &74.29 $\pm$ 3.39 &73.06 $\pm$ 2.83\\
        	                         & progress &  - 1.55 & -1.98 & +0.09 & -4.79 \\\hline
	\multirow{3}{*}{OA3L algorithm} & Init. Acc. & 61.37 $\pm$ 4.87 & 91.92 $\pm$ 6.80& 75.94 $\pm$ 5.17& 79.64 $\pm$ 5.10\\
            	                     & Test Acc. & 55.5 $\pm$ 5.91 &90.11 $\pm$ 7.05 &71.04 $\pm$ 5.58 &72.17 $\pm$ 4.92\\
                	                 & progress & - 5.87 & -1.81 & -4.90 & -7.47 \\
		
	\bottomrule
	\end{tabularx}
	
\end{center}
\end{table*}

In second accuracy test, accuracy of classifier will be compared in two situations. In first situation, all data points will have their corresponding labels which is reported as full budget in Table \ref{tab:acc2}. In second situation, $50\%$ of data points with their labels are used as initialization data and other $50\%$ are used as test data with no labels provided unless 90 labels can be queried per each 1000 data points.

\begin{table*}[h]
\begin{center}
	\captionof{table}{\label{tab:acc2} Accuracy results on comparing full budget and limited budget on test data}
	\begin{tabularx}{0.80\textwidth}{ c | c | c | c | c | c } 
	\toprule
	 & Dataset & PPG & WESAD & Sleep & PAMAP \\\hline
	 & Full budget & 69.39 $\pm$ 1.04 & 98.38 $\pm$ 0.84 & 78.07 $\pm$ 0.91 & 83.57 $\pm$ 1.01\\
	\midrule
	\multirow{2}{*}{Proposed algorithm Linear} & limited budget  & 67.24 $\pm$ 1.83 & 94.16 $\pm$ 1.36& 74.86 $\pm$ 1.88 & 79.98 $\pm$ 2.53\\
                            				 & progress  & - 2.15 & \textbf{-4.22} & \textbf{-3.21} & \textbf{-3.59} \\\hline
    \multirow{2}{*}{Proposed algorithm Exponential} & limited budget & 68.47 $\pm$ 1.39& 93.15 $\pm$ 1.89& 72.95 $\pm$ 2.21& 79.83 $\pm$ 2.61\\
                            				     & progress & \textbf{- 0.92 }& -5.23 & -5.12 & -3.74 \\\hline
	\multirow{2}{*}{winnow algorithm} & limited budget & 62.65 $\pm$ 2.06 &93.29 $\pm$ 3.31 &74.29 $\pm$ 3.39 &73.06 $\pm$ 2.83\\
        	                         & progress &  -6.74  & -5.09 & -3.78 & -10.51 \\\hline
	\multirow{2}{*}{OA3L algorithm} & limited budget & 55.5 $\pm$ 5.91 &90.11 $\pm$ 7.05 &71.04 $\pm$ 5.58 &72.17 $\pm$ 4.92\\
                	                 & progress & - 13.89 & -8.26 & -7.03 & -11.40 \\
		
	\bottomrule
	\end{tabularx}
	
\end{center}
\end{table*}

\par
Confirming the results on quality of labels, Tab. \ref{tab:acc1} shows that the apt query requests causes better performance in classifier. Comparing the exponential and linear proposed algorithms, exponential strategy has the upper hand in three datasets while linear strategy has better performance in one dataset. Results of Tab. \ref{tab:acc2} show that the proposed algorithms have caused better accuracy compared to OA3L and winnow selective sampling. However, results of Tab. \ref{tab:acc2} show that linear strategy has better performance in three datasets and the exponential strategy is better in one dataset. Also, in Tab. \ref{tab:acc1} and Tab.\ref{tab:acc2}, the proposed algorithm with linear and exponential strategies have lower variance compared to OA3L and winnow.

\begin{section}{Discussion}
Results of the experiments show that neural networks are a specific solution for time series label prediction. Neural networks do not consider relativity of time, thus, they are not general optimizer for pattern recognition in time series. Also, the proposed selective sampling algorithm solves selective sampling problem from a different point of view which makes if more reliable and stable by omitting randomness from the algorithm and making decision considering similarity of input stream and \emph{governing pattern}.
\par
Considering concept of \textbf{“time dilation”} derived from the "General Relativity" of Einstein \cite{einstein1915zusammenfassung}, time should be assumed as a milieu of dilating patterns. Because that a neural network classifier has a static architecture, relativity can not be implemented with current architectures. In a classifier which predicts labels for time series, the behavior of time should be justified in "General Relativity" theorem. Hence, any pattern should be predictable when \textbf{“time dilation”} occurs. Despite the implementation details of the proposed algorithm, the stated key difference between the proposed algorithm and neural networks(\emph{Connectionist Architecture}) is what this paper aims to demonstrate.
\end{section}
\begin{section}{Conclusion}
This paper proposes a classifier with particular characteristics for time-series label prediction and also a deterministic selective sampling algorithm which uses the same computational steps as the classifier. The proposed classifier is able to handle variant time spans between data points. Furthermore, the proposed algorithm is able to handle class imbalanced data with no further constraint or computation. Comparing the results of the classifiers, the proposed classifier outperform other time-series prediction classifiers in case of varying sampling rate. Considering that all neural networks are bounded to the initial architecture design, they can be assumed as a special case in time series prediction.
\par
The proposed selective sampling algorithm has better performance in finding the data points that have been predicted in wrong class. In this paper, instead defining a decision line for query strategy, a similarity-based algorithm is proposed to have both classifier and the query strategy working on same principles. Also, the lack of randomness in the proposed selective sampling algorithm, and thus lower variance in reported results, improves the reliability of algorithm in intensive risk applications.
\par
Analysis of online learning step shows that this step is not functioning as expected and neural networks are better learners in case of online learning. Also, online learning in the proposed algorithm is consumes more computational power than neural networks and thus more processing time. In general, results of the experiments show that the proposed algorithm has higher processing time in all situations. Furthermore, when a new feature is added to the input data, the processing time of the proposed algorithm increases faster than neural networks.
\par
This paper proposes an algorithms that preserves all characteristics of the time and thus is able to handle any variant time spans between data points. Also, Considering the differences between \emph{Empiricism} and the \emph{Piagetian oeuvre}, besides counting wrong and correct predictions, the \emph{Matching} step creates an opportunity to investigate \emph{why} the label is predicted in this way. This point of view has weaknesses but paves the way which features new opportunities in case of time series prediction. The proposed sampling strategy is an example of the potential of this point of view. 
\end{section}


%
%


\printbibliography

%
%

\end{document}